# Classifying Text-Based Conspiracy Tweets related to COVID-19 using Contextualized Word Embeddings

Abdul Rehman, Rabeeh Ayaz Abbasi*, Irfan ul Haq Qureshi, Akmal Saeed Khattak

*Department of Computer Science, Quaid-i-Azam University, Islamabad, Pakistan*

**Abstract**
The FakeNews task in MediaEval 2022 investigates the challenge of finding accurate and high-performance models for the classification of conspiracy tweets related to COVID-19. In this paper, we used BERT, ELMO, and their combination for feature extraction and RandomForest as classifier. The results show that ELMO performs slightly better than BERT, however their combination at feature level reduces the performance.

## 1 INTRODUCTION

People widely use social media to share their opinions. At times, people intentionally or unintentionally share false information about serious issues like pandemics. This results in misleading other people about the issue. As a result, people often create hurdles in managing the issue by not complying with the advice provided by authentic sources like healthcare institutes. COVID-19 has affected the entire world [1]. A lot of fake news was spread on social media platforms about the pandemic. Text-Based Misinformation and Conspiracies Detection, one of the subtasks of FakeNews Detection task in MediaEval2022 [2], focuses on the classification of tweets by conspiracy. For this study, we focused solely on this subtask because it was the most closely related to our research question, which was to investigate the effectiveness of contextualized word embeddings for identifying conspiracy-related content in social media. The task uses three different class labels to mark the tweet contents: Promotes/Supports Conspiracy, Discusses Conspiracy, and Non-Conspiracy. There are overall nine conspiracies for which the class labels have to be assigned. We use pre-trained contextualized word embedding models, BERT, and ELMo [3], and their combination for classifying misinformative tweets.

The FakeNews task in 2022 extends from the FakeNews task in MediaEval 2021 [5]. Among its participants, two teams used the contextualized word embeddings from BERT pre-trained model to classify the tweets [6], [7]. In both cases, using the BERT model improved classification performance for different conspiracies.

## 2 METHODOLOGY

In this section, we present the proposed model for the task.

### 2.1 Pre-Processing

We used a tweets dataset from the organizers of the task. Since we are using pre-trained language models like BERT [8] and ELMo [9], preprocessing is not needed. These models use all of the information in a sentence, including punctuation, and stop-words, from a wide range of perspectives by leveraging bidirectional LSTMs and multi-head self-attention mechanisms [10]. As the provided dataset was imbalanced (91% Non-Conspiracy, 3% Discusses Conspiracy, 6% Promotes/Supports Conspiracy), to effectively train the machine learning model, we use "Synthetic Minority Oversampling Technique" (SMOTE). It is a widely used and successful oversampling method to handle unbalanced datasets [11]. We sample the tweets in the following manner: 50% examples of "Non-Conspiracy", 25% "Promotes/Supports Conspiracy", and 25% "Discusses Conspiracy" examples.

### 2.2 Language Models

We compare the performance of pre-trained language models "Bidirectional Encoder Representations from Transformers" (BERT) and "Embeddings from Language Model" (ELMo) for text classification [12], [13]. BERT uses transformers to learn the contextual relationships between words/strings and substrings. The transformer's basic design constitutes an encoder for reading the text input and a decoder for generating the task prediction. The encoder of BERT is sufficient for language modeling and correspondingly for the classification. Like BERT, ELMo also contextualizes word representations. It simulates both the complexity of word use (including syntax and semantics) and the variation in word use.
Usually, large text corpora are used to train these language models using deep bidirectional language model (biLM) functions. They are simple to incorporate into ML models/pipelines and perform good on variety of difficult NLP problems, such as question answering, sentiment analysis, and textual entailment. In addition, we also tried to combine the embeddings from both of these most widely used pre-trained models to get contextualized word embeddings.

The BERT encoder generates a sequence of hidden states for each input sentence, representing the meaning and context of the words in the sentence. To use this sequence for classification tasks, we need to compact it into a single vector representation. There are several ways to do this, such as using max/mean pooling or applying attention to the hidden states. In our study, we chose a simpler approach: we only took the hidden state associated with the initial token ([CLS] token), which is prepended to the beginning of each sentence by BERT as a start-of-sentence token. The representation of the whole tweet is then given by the 768-dimensional embedding of the CLS token, which captures the meaning and context of the entire sentence [14], [15]. Figure 1 shows how BERT embeddings are used for modeling tweets.

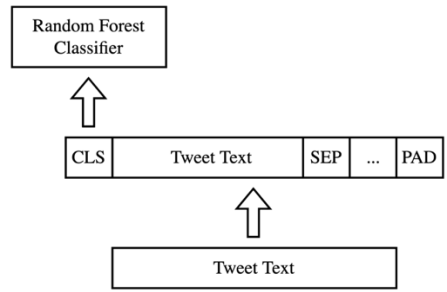

Figure 1: **Usage of BERT Embeddings.**

ELMo gives a vector of size 1024, for every word in the input sentence. To get the vector representation of a tweet as a whole, we use the mean of the ELMo vectors [16]. Figure 2 shows the process of using tweet for classification using ELMO.

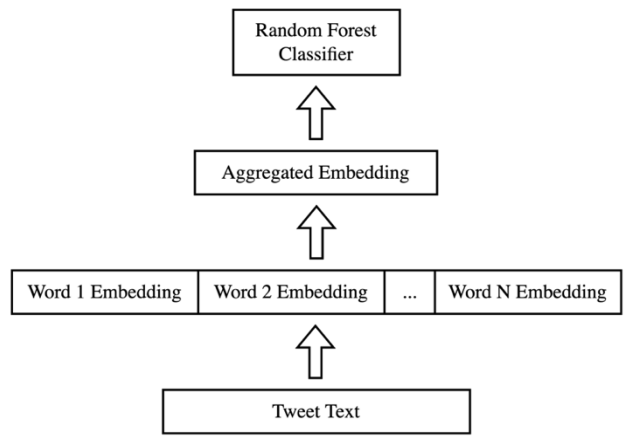

Figure 2: **Usage of ELMo Embeddings.**

Combining the representations of BERT and ELMo is a nontrivial task due to the difference in their dimensionalities. We can reduce the dimensions of the representations using different dimensionality reduction techniques, but it will result in information loss.

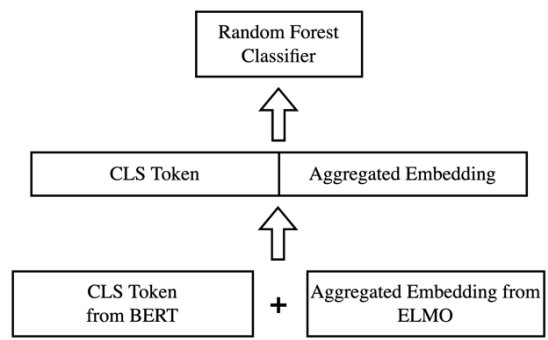

Figure 3: **Usage of Combined BERT-ELMO Embeddings.**

In our study, we combined the representations of BERT and ELMo by concatenating the 768 dimensions of BERT with the 1024 dimensions of ELMo, resulting in a combined representation with 1792 dimensions. This allowed us to take advantage of the strengths of both embeddings without losing any information [17]. Figure 3 shows the combination of BERT and ELMO embeddings.

## 2.3 Classification

In this task, there were nine types of conspiracies and each of them has three labels, and a tweet can belong to more than one conspiracy. It cannot be a problem of multiclass multilabel classification straightaway as the labels are not binary, however, it can be transformed into it. But we used nine different models for each conspiracy and dealt with them separately as a multiclass classification problem [18].

The training dataset constituted 1912 examples. Algorithms used for deep learning typically require large datasets to produce good results, and they often underperform on small datasets [19]. Therefore, we chose to use the Random Forest classifier because it is a robust and widely-used machine learning algorithm that has been shown to perform well on a variety of tasks, including text classification [20]. Random Forest uses multiple decision trees based on subsamples of the dataset [18], [21].

## 3 RESULTS AND ANALYSIS

While applying machine learning algorithms with skewed datasets, traditional error metrics, such as accuracy, are not useful, as they also consider true negatives (or the major class). We use the F1 score, which computes the harmonic mean of precision and recall. Equations 1, 2 and 3 express the formulae of precision, recall and F1 score respectively.

$$Percision = \frac{True\ Positives}{True\ Positives + False\ Positives} \quad (1)$$

$$Recall = \frac{True\ Positives}{True\ Positives + False\ Negatives} \quad (2)$$

$$F_1 Score = \frac{2 \times P \times R}{P + R} \quad (3)$$

Another metric that measures machine learning algorithms' performance for classification problems even if the data is imbalanced is Matthews Correlation Coefficient (MCC). Its value lies between -1 and +1, where -1 represents inverse, 0 represents average, +1 represents perfect predictions. Equation 4 expresses the formula of MCC.

$$MCC = \frac{TP * TN - FP * FN}{\sqrt{(TP + FP) * (TP + FN) * (TN + FP) * (TN + FN)}} \quad (4)$$

For training the model, we split the data into random train and test subsets using 80-20 ratio. Table 1 shows the F1-scores, while Table 2 shows the Matthews Correlation Coefficient (MCC) values of the language models used along with the Random Forest classifier when tested on test subset Table 3 shows the Matthews Correlation Coefficient values of the test data provided by the organizers which also is the official score.

Table 1: **F1 Measures of models using BERT, ELMo, and Combined BERT-ELMo**

| Conspiracy | BERT | ELMo | Combined BERT-ELMo |
|---|---|---|---|
| Suppressed Cures | 0.97 | 0.97 | 0.97 |
| Behavior and Mind Control | 0.84 | 0.85 | 0.83 |
| Antivax | 0.80 | 0.86 | 0.80 |
| Fake Virus | 0.83 | 0.83 | 0.83 |
| Intentional Pandemic | 0.79 | 0.78 | 0.78 |
| Harmful Radiation Influence | 0.94 | 0.94 | 0.94 |
| Population Reduction Control | 0.85 | 0.85 | 0.85 |
| New World Order | 0.88 | 0.87 | 0.89 |
| Satanism | 0.94 | 0.94 | 0.94 |
| **Average F1-Score** | **0.871** | **0.877** | **0.870** |

Table 2: **MCC Values of models using BERT, ELMo, and Combined BERT-ELMo**

| Conspiracy | BERT | ELMo | Combined BERT-ELMo |
|---|---|---|---|
| Suppressed Cures | 0.357 | 0.357 | 0.357 |
| Behavior and Mind Control | 0.145 | 0.149 | 0.000 |
| Antivax | 0.094 | 0.112 | 0.148 |
| Fake Virus | 0.172 | 0.193 | 0.192 |
| Intentional Pandemic | 0.199 | 0.199 | 0.232 |
| Harmful Radiation Influence | 0.000 | 0.000 | 0.000 |
| Population Reduction Control | 0.250 | 0.250 | 0.250 |
| New World Order | 0.210 | 0.161 | 0.236 |
| Satanism | 0.000 | 0.000 | 0.000 |
| **Average MCC Score** | **0.158** | **0.158** | **0.157** |

Table 3: **Detailed Result using MCC of Test Data**

| Conspiracy | BERT | ELMo | Combined BERT-ELMo |
|---|---|---|---|
| Suppressed Cures | -0.007 | -0.007 | -0.007 |
| Behavior and Mind Control | 0.017 | 0.099 | 0.074 |
| Antivax | 0.135 | 0.255 | 0.118 |
| Fake Virus | 0.055 | 0.023 | 0.062 |
| Intentional Pandemic | 0.082 | 0.060 | 0.048 |
| Harmful Radiation Influence | -0.010 | -0.010 | -0.010 |
| Population Reduction Control | 0.011 | -0.017 | -0.017 |
| New World Order | 0.205 | 0.112 | 0.121 |
| Satanism | -0.012 | -0.012 | -0.012 |
| **Average MCC Score** | **0.053** | **0.056** | **0.042** |

ELMo yields better results in most of the cases and even in the remaining cases the differences are small. According to the experimental result, the pre-trained language model aids in the extraction of conspiracy information for stance classification and conspiracy detection. However, in the classification for the Suppressed Cures Conspiracy, Harmful Radiation Influence, and Satanism Conspiracy, all outputs are incorrect. We guess that the classification models do not work because the tweets belonging to other labels (promotes conspiracy or discusses conspiracy) are only a few. Table 4 shows the distribution of training examples across conspiracies.

Table 4: **Percentages of Training Examples across Conspiracies**

| Conspiracy | Non-Conspiracy | Discusses Conspiracy | Promotes / Supports Conspiracy |
|---|---|---|---|
| Suppressed Cures | 97.4% | 2.1% | 0.5% |
| Behavior and Mind Control | 90.8% | 4.9% | 4.3% |
| Antivax | 88.1% | 6.1% | 5.8% |
| Fake Virus | 85.3% | 9.5% | 5.3% |
| Intentional Pandemic | 84.0% | 12.3% | 3.7% |
| Harmful Radiation Influence | 96.5% | 2.0% | 1.5% |
| Population Reduction Control | 91.6% | 6.7% | 1.7% |
| New World Order | 91.0% | 7.8% | 1.2% |
| Satanism | 95.2% | 2.9% | 1.9% |

## 4 CONCLUSIONS

We propose use of different language models like BERT and ELMo, combined with Random Forest Classifier for conspiracy tweets classification. Best results are obtained using the ELMo pretrained model. It achieves an average score of 0.05 MCC without using any augmentation or extra information.